\documentclass[twocolumn]{article}

\usepackage[final]{graphicx}
\usepackage[utf8]{inputenc}
\usepackage[T1]{fontenc}
\usepackage{geometry}
\usepackage{amsmath, amssymb, amsthm}
\usepackage{booktabs}
\usepackage{url}
\usepackage{hyperref}
\hypersetup{colorlinks=true, linkcolor=blue, urlcolor=cyan, citecolor=green}
\usepackage{xcolor}

\usepackage{natbib}
\setcitestyle{numbers,square}

\title{\textbf{DensFiLM: Density-Conditioned Video Saliency\\
for Crowd Scenes}}
\author{%
  \textbf{Anis Ur Rahman}\\
  CSC - IT Center for Science Ltd., Espoo, Finland\\
  \texttt{anis.rahman@csc.fi}
}
\date{}

\newcommand{\mypar}[1]{\noindent\textbf{#1}}

\usepackage{xspace}

\usepackage{booktabs}
\usepackage{tabularx}
\usepackage{caption}
\usepackage{subcaption}

\usepackage{amsmath}
\usepackage{amsfonts}

\newcommand{\ourmodel}{\textsc{DensFiLM}\xspace}

\begin{document}

\twocolumn[
  \begin{center}
    \maketitle
    \thispagestyle{empty}
    \begin{minipage}{0.92\linewidth}
\begin{abstract}
Video saliency models typically apply a single fixation strategy across
crowd scenes, despite systematic changes in attention with crowd density.
Sparse scenes encourage tracking individuals, whereas dense scenes shift
attention toward collective motion and scene-level landmarks. We introduce
\ourmodel, a density-conditioned video saliency model that inserts a
lightweight Feature-wise Linear Modulation layer at the bottleneck of a
Video Swin Transformer. A learned density embedding produces channel-wise
scale and shift parameters, allowing the decoder to reconstruct saliency
from features selected for each density regime. The module adds only
$\approx$\,100K parameters and can use either CrowdFix density labels or
the model's own density prediction.
On CrowdFix, \ourmodel{} achieves mean NSS\,1.434 and CC\,0.517 over four
seeds, improving over ACLNet by 14.7\% and 14.9\%, respectively, while
predicted-density conditioning matches oracle-label performance.
Ablations show that explicit RAFT optical flow and larger temporal and
social-force extensions provide no further improvement in this setting.
In a centre-prior-subtraction diagnostic, density conditioning yields an
NSS gain of $0.462$ over the unconditioned backbone, compared with $0.124$
under standard evaluation. These results show that lightweight bottleneck
conditioning provides a more effective inductive bias than increasing model
capacity for crowd-video saliency.
Our code is available at
\url{https://github.com/aniskhan25/crowdfix-saliency}.
      \end{abstract}
    \end{minipage}
    \vspace{0.3in}
  \end{center}
]

\section{Introduction}%
\label{sec:intro}

Video saliency prediction constructs a computational model of human visual attention over time, estimating where observers fixate while watching dynamic scenes. 
Such models have practical applications in perceptually optimised video compression~\citep{hadizadeh2014}, region-of-interest-guided adaptive streaming, UAV-based crowd analytics, and human-robot interaction in public spaces.
While the field has made rapid progress on generic video benchmarks (Hollywood-2, UCF-Sports, DHF1K), these benchmarks span heterogeneous content (sports, films, surveillance) and the models trained on them implicitly average across fixation strategies that may be mutually incompatible.
Crowd videos present a particularly sharp test case: they belong to a single domain, yet they encompass categorically different attention demands that change with crowd density.

The core hypothesis is what we term the \emph{density-driven fixation shift}.
In \emph{sparse pedestrian} scenes, distinguishable people and trajectories
provide plausible individual fixation targets.  In the \emph{dense
free-flowing} regime, overlap and occlusion reduce individual trackability,
making group motion, flow lanes, and merge regions more informative.  In
\emph{dense congested} scenes, suppressed motion and visual similarity may
increase the importance of scene-level landmarks and crowd boundaries.
These regimes therefore emphasise different visual features.  A single
density-agnostic model must learn all of them from a limited dataset without
an explicit signal indicating which regime is present.

The CrowdFix benchmark~\citep{crowdfix} provides a controlled setting for
studying this phenomenon: 434 crowd clips are annotated with per-frame
fixation maps from 26 eye-tracked observers and partitioned into Sparse
Pedestrian (SP), Dense Free-flowing (DF), and Dense Congested (DC)
categories.  The original CrowdFix study evaluated SAM, DeepVS, and ACLNet,
with ACLNet~\citep{aclnet} obtaining the strongest reported result
(AUC-J\,0.817)~\citep{crowdfix}.  To our knowledge, later video saliency
methods have not reported results on CrowdFix.  Moreover, ACLNet is a
generic attentive CNN-LSTM model and uses the same parameters for every
crowd category, leaving the dataset's density annotations unused.  We argue
that density should instead be introduced at the encoder bottleneck, where
the feature representation is maximally compressed
and semantically abstract.  Modulating the bottleneck controls which
categorical features including individual body silhouettes, mesoscopic flow
trajectories, or scene-landmark contrasts are amplified or suppressed
before decoding begins, thus shaping every subsequent spatial reconstruction
step.

Feature-wise Linear Modulation~\citep{film} provides a suitable
mechanism: an element-wise affine transformation $\tilde{\mathbf{f}} =
\mathbf{f}(1+\gamma) + \beta$, parameterised by a lightweight projection of
a conditioning embedding.  Two properties make FiLM particularly suitable
here.  First, its residual initialisation ($\gamma\!\leftarrow\!0$,
$\beta\!\leftarrow\!0$) ensures that at the start of training the
transformation is an identity, so Kinetics-400-pretrained bottleneck
representations are not disrupted when FiLM is inserted.  Second, per-channel
affine modulation has $O(2C)$ parameter cost for the conditioning projection,
two orders of magnitude cheaper than spatial cross-attention ($O(C^2)$), and
does not require spatial alignment between the conditioning signal and the
feature map, which is appropriate because density is a global scene property, not a
spatially localised one.

To delineate the benefit of targeted conditioning against generic capacity
scaling, we evaluate three progressively complex extensions of the FiLM
baseline: (i) augmenting the model with dense optical flow (RAFT~\citep{raft})
as a parallel motion encoder, fused at the bottleneck via cross-attention;
(ii) a three-branch decoder that adds a parallel UpBlock3d temporal pathway
with a per-category learned social force prior; and (iii) the combination of
both.  All three variants fail to improve over \ourmodel, and the
three-branch architecture, which adds the most parameters, achieves NSS
1.313, returning to the unconditioned VideoSwin baseline value to
three decimal places.  This result is consistent with a bias-variance
tradeoff in which the additional parameters introduce enough estimation
variance to offset the benefit of the architectural priors.  With 3,189
training clips, the tested higher-capacity extensions are less effective
than a minimal, targeted inductive bias.

Our contributions are:
\begin{enumerate}
  \item \textbf{Density-conditioned bottleneck modulation.}
        We introduce \ourmodel, which conditions semantically abstract
        Video Swin features on crowd density using a lightweight FiLM layer.
        A one-hot ablation shows that bottleneck injection provides most of
        the improvement, while feature-wise scaling offers additional
        class-specific recalibration at a cost of only $\approx$\,100K
        parameters.
  \item \textbf{Best reported CrowdFix results and practical inference.}
        \ourmodel{} achieves AUC-J\,0.820, NSS\,1.434, and CC\,0.517
        over four seeds on CrowdFix.  Conditioning on the model's own
        density predictions matches oracle-label performance within
        $0.001$ NSS, removing the need for manual density labels at inference.
  \item \textbf{Centre-prior and capacity-aware analysis.}
        Subtracting the mean training fixation density increases the measured
        conditioning gain from $0.124$ to $0.462$ NSS\@.  Ablations show that explicit RAFT
        optical flow is redundant and that larger temporal and social-force
        variants fail to improve, indicating that the available CrowdFix
        training data do not support the additional capacity tested here.
\end{enumerate}

\section{Related Work}%
\label{sec:related}

\subsection{Video Saliency Prediction}

Saliency modelling originated with biologically motivated centre-surround
feature maps combining multi-scale luminance, colour, and motion
contrasts~\citep{itti1998}.  Deep learning approaches replaced these
hand-crafted features with learned representations.
SAM~\citep{sam} augmented spatial saliency with a
multi-level attentive LSTM that integrates CNN features across resolution
scales; applied to video, it generates per-frame maps without explicit
inter-frame temporal modelling.  DeepVS~\citep{deepvs}
introduced a two-stream architecture combining appearance CNN features with an
optical flow stream, fused by a temporal LSTM for short-range motion
integration; while effective for activities with salient individual motion,
it lacks the capacity for long-range temporal structure and operates with a
VGG-level backbone that limits representational power.  TASED-Net~\citep{tased}
replaced recurrent integration entirely with a
fully 3D S3D~\citep{s3d} encoder pretrained on Kinetics-400~\citep{kinetics},
enabling rich temporal attention at the feature level.  TMFI-Net~\citep{tmfi_net}
uses a Video Swin backbone, a semantic-guided encoder, and a hierarchical
decoder to integrate multi-scale spatiotemporal features.  SalFoM~\citep{salfom}
demonstrated that large-scale video foundation model pretraining
substantially improves saliency generalisation when task-specifically adapted.
GASP~\citep{gasp} incorporates social cues, including gaze direction and
affect, through gated late-fusion mechanisms.

None of these methods use crowd density as a conditioning signal, and their
original publications do not report CrowdFix results.  Our published
baseline comparison therefore uses SAM, DeepVS, and ACLNet from the
original CrowdFix study~\citep{crowdfix}; we additionally train TASED-Net
on the same split to provide a stronger modern baseline.

Our work uses the Video Swin Transformer~\citep{video_swin}, whose
hierarchical shifted 3D window attention provides multi-scale
spatiotemporal representations.

\subsection{Crowd-Specific Video Saliency}

The CrowdFix study~\citep{crowdfix} established a benchmark by evaluating
three existing saliency models.  ACLNet~\citep{aclnet}, the strongest of
these baselines, combines attention-enhanced convolutional features with
recurrent temporal modelling, but it was designed for generic video saliency
and does not use CrowdFix density categories.  Consequently, all three
categories are processed by the same representation and decoder.

In contrast, \ourmodel{} conditions the encoder bottleneck using a coarse
3-class density label provided by the dataset annotation.  At inference,
this label can be replaced by the model's auxiliary density prediction,
requiring no external density-estimation network.  This design targets the
dataset's central structure directly: the density category controls the
abstract feature channels before spatial decoding begins.

\subsection{Feature-wise Linear Modulation}

FiLM~\citep{film} was introduced in the context
of visual question answering, where it was shown that language-derived affine
parameters could steer a ResNet's internal feature representations toward
question-relevant visual attributes, outperforming both feature concatenation
and attention-based conditioning.  The element-wise affine form
$\tilde{\mathbf{f}} = (1+\gamma(\mathbf{c})) \odot \mathbf{f} + \beta(\mathbf{c})$
allows selective amplification of channels that are diagnostic for the
current conditioning context while suppressing those that are not, without
altering the spatial structure of the feature map.  A key practical
advantage is its residual initialisation: with
$\gamma\!\leftarrow\!0$ and $\beta\!\leftarrow\!0$, FiLM begins as an
identity transformation, enabling stable integration into a pretrained
architecture without disrupting learned representations at the start of
fine-tuning.  Our application of FiLM to crowd density conditioning in video
saliency is, to our knowledge, the first in this domain.

\subsection{Crowd Density Estimation and \mbox{Dynamics}}

Crowd density estimation~\citep{zhang2016} typically predicts local
pedestrian density as a Gaussian-blurred sum of annotated head positions,
used for crowd counting and management.  The Social Force
Model~\citep{helbing} describes pedestrian dynamics as emergent behaviour
arising from pairwise repulsive/attractive force fields between individuals,
goals, and boundaries, predicting lane formation, avoidance patterns, and
stop-and-go waves at the collective level.  These models motivate our
three-branch ablation variant (Section~\ref{sec:ablation}), which implements
a learnable spatial social force prior parameterised per density category.
This extension did not improve saliency performance in our experiments,
suggesting that its additional structure is not beneficial at the available
training scale.

\section{CrowdFix Dataset}%
\label{sec:dataset}

CrowdFix~\citep{crowdfix} contains 434 short clips derived from 89 source
videos at $1280\!\times\!720$ resolution and 30\,fps, covering real-world
crowd environments including
gymnasiums, transit concourses, public squares, and subway platforms.
Eye-tracking data were collected from 32 participants using a calibrated
remote eye tracker, with 26 participants retained after data-quality
filtering.  Ground-truth saliency maps are constructed by convolving the
binary fixation maps with a Gaussian kernel corresponding to
$1^\circ$ of visual angle ($\sigma\!=\!38$\,px); the binary maps are retained
for fixation-based metrics.

The dataset is partitioned into three density categories:
\begin{itemize}
  \item \textbf{SP (Sparse Pedestrian):}  116 clips with clearly
        distinguishable individual pedestrians exhibiting unoccluded motion
        trajectories (gymnasiums, pedestrian crossings, open plazas).
  \item \textbf{DF (Dense Free-flowing):}  163 clips with high pedestrian
        density but unrestricted directional flow (transit concourses,
        open-air markets, stadium plazas at capacity).
  \item \textbf{DC (Dense Congested):}  155 clips with extreme density
        and restricted individual motion (subway platforms, narrow passageways,
        protest gatherings).
\end{itemize}

We adopt a 70/15/15 train/validation/test split stratified by category.
Eight-frame windows are extracted with stride~4 and resized to
$224\!\times\!384$ pixels, yielding 3,189 training windows, 667 validation
windows, and 5,356 evaluated frames in the test split.  This training set
is substantially smaller than generic video saliency benchmarks
(DHF1K: 1,000 full videos; Hollywood-2: 1,707 clips), making model
regularisation a dominant design consideration rather than a secondary concern.

\section{DensFiLM}%
\label{sec:method}

\begin{figure*}[!ht]
    \centering
    \includegraphics[width=0.96\linewidth]{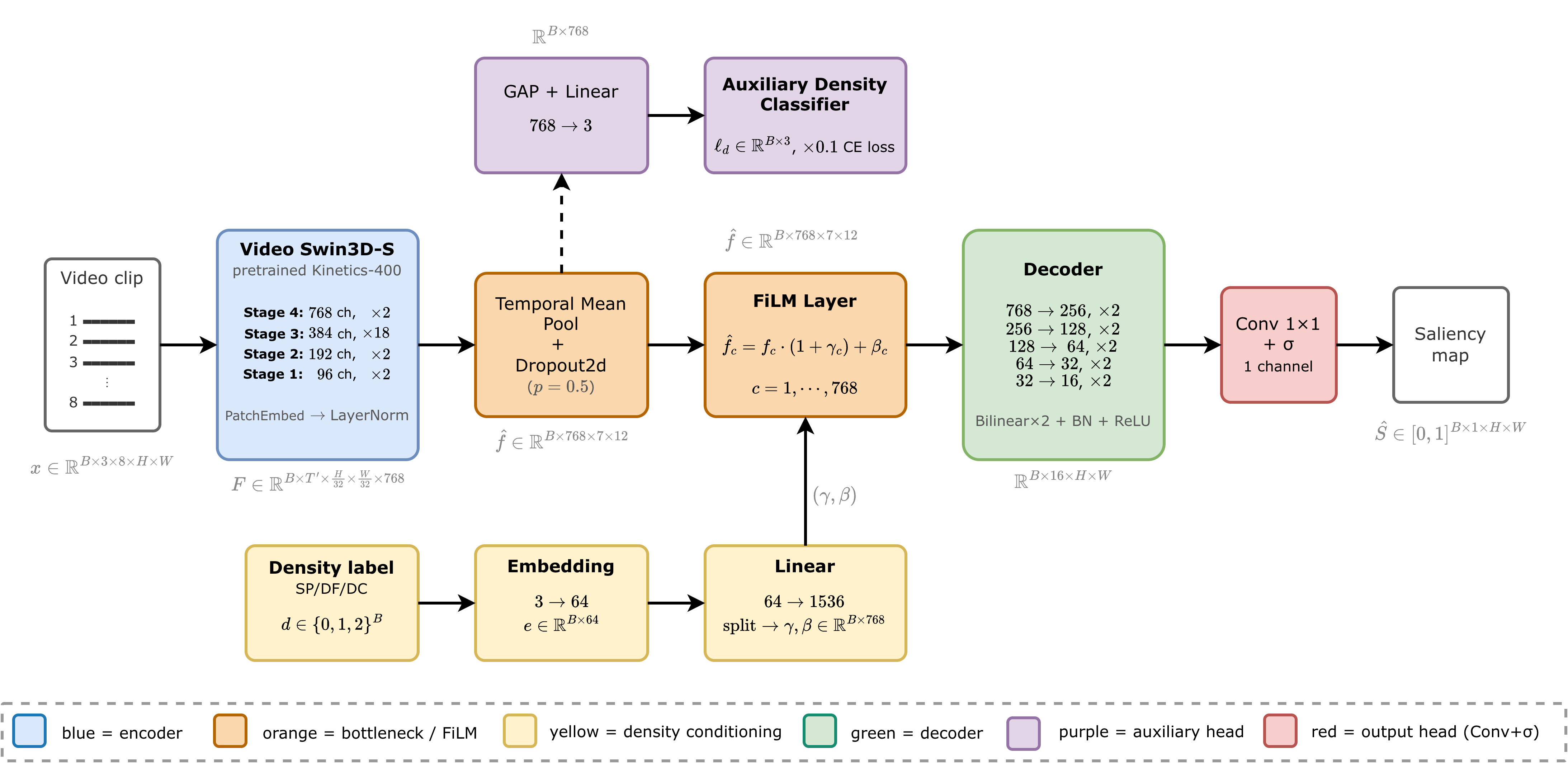}
    \caption{%
      Overview of \ourmodel.  An 8-frame crowd video clip is processed by a
      Video Swin-S encoder, producing a spatiotemporal feature volume
      $\mathbf{F}\in\mathbb{R}^{B\times4\times7\times12\times768}$.  Temporal
      mean pooling and Dropout2d ($p\!=\!0.5$) produce a 2D bottleneck
      $\mathbf{f}\in\mathbb{R}^{B\times768\times7\times12}$.  A FiLM layer
      conditioned on the 3-class density label embedding applies per-channel
      scale-shift modulation (Eq.~\ref{eq:film}), reshaping which abstract
      features are emphasised before the five-stage 2D decoder reconstructs
      the saliency map at full resolution.  An auxiliary density classification
      head enforces discriminative bottleneck structure.
    }%
    \label{fig:architecture}
\end{figure*}

Figure~\ref{fig:architecture} provides a complete architectural overview of
\ourmodel.  The design philosophy is minimal interventionism: a strong
pretrained spatiotemporal encoder (Swin3D-S on Kinetics-400) is preserved
with targeted conditional modulation at the single architectural location
where it is most effective.

\subsection{Problem Formulation}

Let $\mathbf{X} \in \mathbb{R}^{B \times 3 \times T \times H \times W}$
denote a batch of crowd video clips and $d \in \{0, 1, 2\}$ the coarse
crowd density label (SP\,$=\!0$, DF\,$=\!1$, DC\,$=\!2$).  The goal is to
predict a target-frame spatial saliency map
$\hat{\mathbf{s}} \in \mathbb{R}^{B \times H \times W}$, with values in
$[0,1]$, that approximates the
ground-truth fixation density $\mathbf{s}^{*}$.

\subsection{Video Swin Encoder}

We adopt Video Swin Transformer Small (Swin3D-S)~\citep{video_swin}
pretrained on Kinetics-400~\citep{kinetics} as the spatiotemporal encoder.
The Swin3D-S architecture applies shifted 3D window self-attention across
four hierarchical stages at progressively coarser spatial and temporal
resolutions, with patch-merging operations reducing resolution between stages.
This hierarchical temporal attention is crucial for our design: by the final
encoder stage, the bottleneck features encode multi-scale temporal dynamics
ranging from low-level motion contrast (Stage~1) to high-level action and
scene semantics (Stage~4).  The encoder maps a clip at $(T\!=\!8$,
$H\!=\!224$, $W\!=\!384)$ to:
\begin{align}
  \mathbf{F} = \mathrm{Swin3D\text{-}S}(\mathbf{X})
  \;\in\; \mathbb{R}^{B \times T' \times H' \times W' \times C}
  \label{eq:encoder}
\end{align}
where $T'\!=\!4$, $H'\!=\!7$, $W'\!=\!12$, $C\!=\!768$.

The choice of Swin3D-S over simpler alternatives is deliberate.  Its
temporally-shifted window attention integrates frame-to-frame correspondence
at multiple representational levels, encoding implicit motion trajectories
at the level of semantic tokens rather than raw pixel velocities.  The RAFT
ablation in Section~\ref{sec:ablation} tests whether explicit optical flow
adds information beyond these learned temporal features.

\subsection{Bottleneck Compression}

The spatiotemporal volume is compressed to a 2D spatial feature map by
temporal mean pooling over the $T'\!=\!4$ temporal tokens:
\begin{align}
  \mathbf{f} = \frac{1}{T'}\sum_{t=1}^{T'} \mathbf{F}_{:,t,:,:,:}
  \;\in\; \mathbb{R}^{B \times C \times H' \times W'}
  \label{eq:pool}
\end{align}

Temporal mean pooling compresses the encoder output into a single spatial
feature map while retaining temporally aggregated information in the
channel representation.  This design deliberately trades explicit temporal
resolution in the decoder for a compact bottleneck and is evaluated against
the 3D decoder ablation in Section~\ref{sec:ablation}.

Spatial Dropout~\citep{pytorch} with $p\!=\!0.5$ is applied to the bottleneck
before FiLM conditioning.  Critically, Dropout2d drops entire feature
\emph{channels} (all spatial positions within a channel simultaneously),
as opposed to element-wise dropout that corrupts individual activations.
This channel-structured dropout randomly suppresses complete feature maps
during training, encouraging the decoder not to rely on a small subset of
bottleneck channels.

\subsection{Density-Conditioned Bottleneck via FiLM}%
\label{sec:bottleneck}

A learnable density embedding
$\mathbf{e} = \mathrm{Emb}(d) \in \mathbb{R}^{64}$ maps the scalar
density label to a 64-dimensional representation, which is projected to
FiLM scale-shift parameters:
\begin{align}
  [\gamma;\; \beta] = \mathrm{Linear}_{64 \to 2C}(\mathbf{e}),
  \qquad \gamma,\beta \in \mathbb{R}^{C}
  \label{eq:film_proj}
\end{align}

The modulated bottleneck is formed as:
\begin{align}
  \tilde{\mathbf{f}} = \mathbf{f} \cdot (1 + \gamma) + \beta
  \label{eq:film}
\end{align}
where $\gamma$ and $\beta$ are broadcast over the spatial dimensions
$(H', W')$.  The FiLM projection weights and biases are zero-initialised,
so the residual form $(1+\gamma)$ begins as the identity transformation
($\gamma\!=\!0$, $\beta\!=\!0$).  The frozen-backbone phase therefore
starts from the original Kinetics-400 representations.  As training
progresses, $\gamma$ and $\beta$ diverge from zero to encode
density-specific channel weighting.

We choose the bottleneck as the injection point for three reasons.  First,
it is the most semantically abstract representation in the network:
individual pixels have been aggregated into semantic tokens encoding
motion trajectories, group structures, and scene categories.  Second,
FiLM modulation here propagates through all five decoder stages, so
every spatial scale of the reconstructed saliency map is shaped by the
density context.  Third, early-stage conditioning (at the input or
shallow features) would require density-conditioned low-level feature
extraction, while the relevant distinction between density categories
is semantic, not low-level.

The $\approx$100K parameters of the FiLM module comprise a $3\times64$
embedding matrix, a $64\times1536$ linear projection, and their biases,
representing $<\!0.2$\% of Swin3D-S's $\approx$50M parameters.

We note that with $K\!=\!3$ density classes, the embedding table
degenerates to three fixed $(\gamma,\beta)$ parameter vectors: class
lookup rather than interpolative modulation.  The 64-dim bottleneck does
not buy the continuous interpolation it would provide in a many-class or
continuous conditioning setting.  Section~\ref{sec:film_analysis} shows
empirically, however, that the three learned parameter vectors occupy
nearly orthogonal directions ($\max\text{ pairwise cos}(\gamma_i,\gamma_j)
\!=\!-0.03$), indicating class-specific recalibration rather than a shared
offset.  Section~\ref{sec:ablation} (Table~\ref{tab:ablation})
further includes a one-hot bias baseline that removes the embedding
geometry entirely; the FiLM model's advantage over this baseline
quantifies the contribution of the learned embedding structure.

An auxiliary density classification head consists of global average pooling
of $\mathbf{f}$ followed by $\mathrm{Linear}_{768\to3}$, producing
logits for a 3-class density prediction $\hat{d}$.  This head serves a
dual purpose: it encourages the bottleneck feature distribution to remain
discriminative across density categories, and it supplies the predicted
label used by the non-oracle inference mode.

\subsection{Saliency Decoder}

The decoder reconstructs the saliency map from $\tilde{\mathbf{f}}$ through
five successive UpBlock2d modules, each performing bilinear $2\times$
upsampling followed by a $3\times3$ Conv-BN-ReLU:
\begin{align}
  \mathrm{UpBlock2d}(\mathbf{f}) =
  \mathrm{Conv}_{3\times3}\!\bigl(\mathrm{BN}\bigl(\mathrm{ReLU}(\uparrow_2 \mathbf{f})\bigr)\bigr)
  \label{eq:upblock}
\end{align}

The channel progression $768\!\to\!256\!\to\!128\!\to\!64\!\to\!32\!\to\!16$
achieves a cumulative $32\times$ spatial expansion from $7\!\times\!12$ to
$224\!\times\!384$, exactly matching the input resolution.  The rapid
channel reduction at Stage~1 ($768\to256$) compresses the high-dimensional
bottleneck representation into a more distributed spatial format early in
decoding, concentrating the information bottleneck at the stage where
density conditioning has already shaped which channels are salient.  Each
subsequent stage refines spatial detail at progressively finer scales.  A
final $1\!\times\!1$ convolution followed by sigmoid produces the saliency
map:
\begin{align}
  \hat{\mathbf{s}} = \sigma\!\left(\mathrm{Conv}_{1\times1}
  \!\left(\mathbf{f}^{(5)}_\mathrm{dec}\right)\right)
  \label{eq:sigmoid}
\end{align}

\subsection{Training Objectives}

The primary saliency loss combines KL divergence and Pearson correlation:
\begin{align}
  \mathcal{L}_\mathrm{sal} =
  \mathcal{L}_\mathrm{KL}\!\left(\hat{\mathbf{s}},\mathbf{s}^{*}\right)
  - \lambda_\mathrm{CC}\, \mathrm{CC}(\hat{\mathbf{s}}, \mathbf{s}^{*}),
  \qquad \lambda_\mathrm{CC}\!=\!0.2
  \label{eq:sal_loss}
\end{align}

The KL term measures distributional divergence between the normalised
prediction and ground-truth saliency maps.  The $-\mathrm{CC}$ term provides
a complementary objective that rewards global spatial correspondence.

The total loss incorporates the auxiliary density classification cross-entropy:
\begin{align}
  \mathcal{L} = \mathcal{L}_\mathrm{sal}
  + \lambda_\mathrm{cls}\,\mathcal{L}_\mathrm{cls},
  \qquad \lambda_\mathrm{cls}\!=\!0.1
  \label{eq:total_loss}
\end{align}

The auxiliary weight $\lambda_\mathrm{cls}\!=\!0.1$ is small enough that
the density classifier is a subordinate regulariser rather than a competing
primary objective, but large enough to maintain discriminative structure
in the bottleneck throughout training.

\subsection{Differential Training Protocol}

The Video Swin backbone is frozen for the first 20 epochs while the decoder
and FiLM module receive full gradients, then unfrozen with a $10\times$
reduced learning rate:
\begin{align}
  \eta_\mathrm{backbone} = 0.1 \times \eta_\mathrm{decoder}
  \label{eq:diff_lr}
\end{align}

This two-phase protocol serves a specific purpose.  During the frozen phase,
the decoder learns a useful mapping from Kinetics-400 Swin3D-S features to
CrowdFix saliency patterns, and the FiLM parameters begin to diverge from
zero to encode density-discriminative channel weighting, all without
disturbing the pretrained encoder.  This warmup reduces the risk that early
decoder gradients destabilise the pretrained temporal representations.
The subsequent
fine-tuning phase with $\eta_\mathrm{backbone}\!=\!10^{-5}$ allows the
encoder to adapt at the margins toward CrowdFix-specific motion patterns
without catastrophic forgetting of the temporal representations that make
the backbone effective.

\section{Experiments}%
\label{sec:experiments}

\subsection{Implementation Details}

All models are implemented in PyTorch~\citep{pytorch} and trained with
Distributed Data Parallel across 8 AMD MI250X GCDs on the LUMI HPC cluster.
Training uses Adam~\citep{adam}
with $\eta\!=\!10^{-4}$, weight decay $10^{-4}$, and batch size 2 per GCD
(effective batch size 16).  Training runs for at most 150 epochs with early
stopping on validation NSS (patience\,=\,20).  All ablation variants use
identical optimisation hyperparameters.  Clips of length
$T\!=\!8$ at stride~4 are input at $224\!\times\!384$ resolution with
horizontal flip augmentation during training.  Architectural choices
(backbone, dropout rate, loss weights) were finalised on validation NSS
before test evaluation; all ablation variants are compared on the
held-out test set, with no further selection after that point.

\ourmodel{} supports two evaluation modes: \textbf{oracle}, in which
the ground-truth CrowdFix category label (SP/DF/DC) is provided at
inference; and \textbf{predicted}, in which the model uses its own
auxiliary density head's argmax prediction as the label.  Both modes
are reported in Table~\ref{tab:sota}.

\subsection{Evaluation Metrics}

We report six standard saliency metrics~\citep{mit_benchmark}:
AUC-Judd (AUC-J, $\uparrow$) measures the area under the ROC curve
treating saliency prediction as a fixation-vs-non-fixation classifier,
assessing ranked ordering.  AUC-Borji (AUC-B, $\uparrow$) uses repeated
sampling of non-fixated pixels to estimate the ROC curve.  NSS
($\uparrow$) measures the mean normalised
saliency value at actual fixation locations, assessing calibrated magnitude
at ground-truth fixations.  CC ($\uparrow$) is the Pearson correlation
between predicted and ground-truth saliency, measuring global spatial
co-variation.  KL divergence ($\downarrow$) measures distributional
divergence between the predicted saliency map and the ground truth treated
as probability distributions.  SIM ($\uparrow$) measures the histogram
intersection (Bhattacharyya coefficient) between normalised prediction
and ground truth.  These metrics are complementary: AUC-J assesses ranking,
NSS assesses precision at fixation locations, CC and SIM assess spatial
structure, and KL assesses distribution calibration.

\subsection{Comparison with State of the Art}%
\label{sec:sota}

Table~\ref{tab:sota} reports overall test-set performance for all methods.

\begin{table*}[!t]
  \def\arraystretch{1.1}
  \centering
  \footnotesize
  \caption{%
    Comparison with prior work on the CrowdFix test set (5,356 evaluated frames).
    Published baseline results from~\citep{crowdfix};
    SAM/DeepVS NSS and CC from the same benchmark log.
    ``-'': metric not reported or not applicable.
    \textbf{Bold}: primary multi-seed result.
    Oracle label row reports mean\,$\pm$\,std across 4 random seeds
    (seeds 0, 1, 2, and 42).
    $\dagger$: pred.\ label row is a single-seed oracle-parity check (seed~42).
    $\S$: Centre-prior baseline predicts the mean training fixation density
    for every test sample (no learned model).
    $\ddagger$: Prior-subtracted = mean training fixation density subtracted from model
    predictions before metric computation; KL/SIM omitted as they require
    non-negative distributions.
  }
  \resizebox{\textwidth}{!}{%
  \begin{tabular}{lcccccc}
    \toprule
    \textsc{Model}
      & \textsc{AUC-J}$\uparrow$
      & \textsc{AUC-B}$\uparrow$
      & \textsc{NSS}$\uparrow$
      & \textsc{CC}$\uparrow$
      & \textsc{KL}$\downarrow$
      & \textsc{SIM}$\uparrow$ \\
    \midrule
    SAM~\citep{sam}
      & 0.777 & - & 0.894 & 0.325 & - & - \\
    DeepVS~\citep{deepvs}
      & 0.788 & - & 0.985 & 0.401 & - & - \\
    ACLNet~\citep{aclnet}
      & 0.817 & - & 1.250 & 0.450 & - & - \\
    \midrule
    TASED-Net (ours, trained here)
      & 0.810 & 0.814 & 1.293 & 0.475 & 1.121 & 0.422 \\
    VideoSwin (no FiLM)
      & 0.806 & 0.810 & 1.313
      & 0.476 & 1.133 & 0.424 \\
    \textbf{\ourmodel{} (ours, oracle label)}
      & $\mathbf{0.820}\pm0.001$ & $\mathbf{0.825}\pm0.001$ & $\mathbf{1.434}\pm0.007$
      & $\mathbf{0.517}\pm0.002$ & $\mathbf{1.068}\pm0.004$ & $\mathbf{0.449}\pm0.003$ \\
    \ourmodel{} (pred.\ label)$^\dagger$
      & 0.821 & 0.825 & 1.438
      & 0.518 & 1.067 & 0.451 \\
    \midrule
    \multicolumn{7}{l}{\small\textit{Centre-prior subtraction diagnostic (seed~42)}} \\[2pt]
    Centre-prior baseline$^\S$
      & 0.831 & 0.835 & 1.557 & 0.556 & 0.999 & 0.464 \\
    VideoSwin (prior-subtracted$^\ddagger$)
      & 0.669 & 0.671 & 0.726 & 0.270 & - & - \\
    \ourmodel{} (ours, prior-subtracted$^\ddagger$)
      & 0.765 & 0.769 & 1.188 & 0.431 & - & - \\
    \bottomrule
  \end{tabular}}%
  \label{tab:sota}
\end{table*}

TASED-Net trained from scratch on CrowdFix already surpasses ACLNet on
NSS (1.293 vs.\ 1.250) and CC (0.475 vs.\ 0.450), confirming that
modern temporal modelling alone improves over the strongest baseline from
the original CrowdFix study.  Kinetics-400-pretrained VideoSwin further
improves NSS to 1.313, and density FiLM conditioning then adds the
additional +0.124 NSS step.  The proposed model surpasses ACLNet across all
reported metrics: $\Delta$AUC-J\,=\,+0.003, $\Delta$NSS\,=\,+0.184
(+14.7\%), $\Delta$CC\,=\,+0.067 (+14.9\%).  Several aspects of this
comparison warrant detailed analysis.

\mypar{Asymmetric NSS vs.\ AUC-J gains.}
The most informative comparison is the relative magnitude of NSS and AUC-J
improvements.  FiLM conditioning over the VideoSwin baseline improves NSS
by $+9.2\%$ but AUC-J by only $+1.7\%$.  This asymmetry reflects the
fundamental difference between these metrics: AUC-J is a ranking metric
assessing whether fixation pixels receive higher predicted saliency than
non-fixation pixels. The VideoSwin backbone's temporally-aware features
already rank these pixels reasonably well.  NSS, by contrast, measures
the normalised predicted saliency value \emph{precisely at} the fixation
locations, quantifying how sharply and accurately the model fires at the
exact spatial positions observers fixated.  The large NSS gain with modest
AUC-J gain indicates that FiLM conditioning primarily improves the
\emph{calibrated precision} of saliency responses at fixation locations,
not the gross pixel ranking.  Mechanistically, this is consistent with the
FiLM modulation selectively reweighting channels associated with the type of visual
feature that is actually fixated in each density category: individual
silhouettes in SP, flow structures in DF, and scene landmarks in DC\@. This may produce
sharper, more precisely placed peaks rather than simply reordering which
pixels score highest.  The $-5.8\%$ KL reduction (1.133~$\to$~1.068) and
$+8.6\%$ CC improvement corroborate this interpretation: both metrics are
sensitive to the shape and spatial structure of the saliency distribution,
not merely its ranking.

\mypar{VideoSwin backbone advantage.}
Comparing our VideoSwin baseline (AUC-J\,0.806, NSS\,1.313) against
ACLNet (0.817, 1.250) reveals an interesting crossover: ACLNet achieves
higher AUC-J ranking than our unconditioned baseline, suggesting that its
attention and recurrent temporal modelling provide effective spatial
ranking.  However, its NSS (1.250) is lower than our VideoSwin baseline
(1.313), meaning that despite better ranking, ACLNet's predicted saliency
values are less precisely calibrated at actual fixation locations.
This reflects the representational advantage of Kinetics-400-pretrained
Swin3D-S for temporal motion encoding: the stronger backbone captures
temporally-resolved motion patterns that directly predict where observers
fixate, rather than only which areas are more salient than others in a
coarse sense.  The addition of FiLM conditioning then provides the precision
gains on top of this strong backbone foundation.

\mypar{Centre-prior subtraction diagnostic.}
Standard saliency metrics conflate crowd-content-responsive predictions with
centre-prior alignment: if a model learns to predict a fixed central blob
(matching the dataset's overall fixation tendency), it will score well on
NSS and AUC-J regardless of whether it understands crowd structure.
As a diagnostic of sensitivity to this shared prior, we subtract the mean
training-set fixation density from all model predictions before computing
metrics (Table~\ref{tab:sota}, lower panel).  This is not a standard
benchmark protocol and can produce negative values, so we omit KL and SIM
and interpret the results only as a relative comparison between models.

The centre-prior baseline predicts the mean training fixation density
for every test sample and achieves NSS\,1.557, exceeding both
VideoSwin (1.313) and \ourmodel{} (1.437) in standard evaluation.
This strong baseline shows that centre-prior alignment contributes
substantially to standard CrowdFix scores.

After prior subtraction, VideoSwin's NSS decreases from 1.313 to 0.726
(45\%), whereas \ourmodel{} decreases from 1.437 to 1.188 (17\%).
This difference is consistent with density conditioning reducing sensitivity
to the shared spatial prior.  The resulting \emph{prior-subtracted
conditioning gain} is
$\Delta\mathrm{NSS}\!=\!+0.462$ ($1.188 - 0.726$), which is
$3.7\!\times$ the standard gain of $+0.124$.

This diagnostic suggests that density conditioning improves crowd-responsive
prediction rather than merely sharpening the shared centre prior.

\subsection{Ablation Study}%
\label{sec:ablation}

Table~\ref{tab:ablation} evaluates the contribution of each component.
All variants share the same VideoSwin backbone, 5-stage 2D decoder, and
hyperparameters; they differ only in what is added beyond the baseline.

\begin{table}[!t]
  \def\arraystretch{1.1}
  \centering
  \footnotesize
  \caption{%
    Ablation study on the CrowdFix test set (single seed, seed~42).
    $\Delta$NSS: gain over VideoSwin baseline (standard evaluation).
    $\Delta$NSS$_\text{prior}$: gain over prior-subtracted VideoSwin
    (NSS\,0.726); ``-'' where prior subtraction was not run.
    The ``+Density FiLM'' row reports seed~42 values; Table~\ref{tab:sota}
    reports mean\,$\pm$\,std over 4 seeds.
    The second panel (Soft FiLM; Multiscale FiLM; Continuous) reports Phase~2 variants
    that explore conditioning without GT labels at inference, multi-scale
    injection, and continuous VGG16 features, respectively.
    The bottom panel (Phase~3) reports cross-dataset transfer: VideoSwin
    pretrained on DHF1K then fine-tuned as DensFiLM on CrowdFix.
  }
  \resizebox{\columnwidth}{!}{%
  \begin{tabular}{lccccc}
    \toprule
    \textsc{Variant}
      & \textsc{AUC-J}
      & \textsc{NSS}
      & \textsc{CC}
      & $\Delta$\textsc{NSS}
      & $\Delta$\textsc{NSS}$_\text{prior}$ \\
    \midrule
    VideoSwin (baseline)
      & 0.806 & 1.313 & 0.476 & - & - \\
    +~Density FiLM (\textbf{ours})
      & 0.820 & 1.437 & 0.517 & $+$0.124 & $+$0.462 \\
    \midrule
    +~One-hot density code (no FiLM)
      & 0.819 & 1.424 & 0.514 & $+$0.111 & $+$0.345 \\
    +~Optical flow (RAFT)
      & 0.820 & 1.437 & 0.517 & $+$0.124 & - \\
    +~3D decoder + social prior
      & 0.808 & 1.313 & 0.477 & $\pm$0.000 & - \\
    \midrule
    +~Soft FiLM (predicted density)
      & 0.820 & 1.424 & 0.514 & $+$0.111 & $+$0.446 \\
    +~Multiscale FiLM (strides 32/16/8/4)
      & 0.819 & 1.412 & 0.512 & $+$0.099 & $+$0.405 \\
    +~Continuous (VGG16 features, no GT label)
      & 0.821 & 1.441 & 0.520 & $+$0.128 & $+$0.416 \\
    \midrule
    \multicolumn{6}{l}{\textit{Phase 3: cross-dataset pretraining}} \\
    DHF1K pretrain $\to$ CrowdFix finetune
      & 0.821 & 1.436 & 0.520 & $+$0.123 & - \\
    \bottomrule
  \end{tabular}}%
  \label{tab:ablation}
\end{table}

\mypar{Density FiLM provides the main improvement.}
FiLM bottleneck conditioning improves NSS by $+0.124$ and CC by $+0.041$
over the plain VideoSwin baseline, with no other architectural changes.
The gain over the unconditioned VideoSwin backbone is attributable to a
single $64\!\times\!1536$ linear projection and a $3\!\times\!64$ embedding
matrix.  The result shows that the density label supplies useful information
beyond the unconditioned video features.

\mypar{One-hot conditioning ablation.}
To distinguish the contribution of the FiLM mechanism from the benefit of
having any density signal, we train a variant that injects a 3-class
one-hot vector via a direct $3\!\times\!768$ linear bias added to the
bottleneck, with no learned embedding and no multiplicative $\gamma$ pathway.
This variant, \textit{+One-hot density code}, uses the same backbone,
decoder, auxiliary classification head, and hyperparameters as \ourmodel{}
but replaces the $64$-dim embedding and FiLM projection with the 3-dim
one-hot bias (2,304 parameters vs.\ 100K for FiLM).

The one-hot model achieves NSS\,1.424, AUC-J\,0.819, CC\,0.514
(Table~\ref{tab:ablation}), versus \ourmodel's NSS\,1.437, AUC-J\,0.820,
CC\,0.517 (seed~42).  Two findings follow from standard evaluation.
First, the conditioning \emph{signal} is the primary driver: the one-hot model
captures $\Delta$NSS\,=\,$+0.111$ of FiLM's $+0.124$ total gain (89\%).
Second, the FiLM \emph{mechanism} adds $+0.013$ NSS\@.  This difference is
small relative to the observed seed variability and should therefore be
interpreted cautiously.

The centre-prior-subtracted diagnostic further separates the variants
(Table~\ref{tab:ablation}, $\Delta$NSS$_\text{prior}$ column).
The one-hot model's prior-subtracted NSS is 1.071
($\Delta$NSS$_\text{prior}$\,=\,$+0.345$ over VideoSwin 0.726), while
\ourmodel{} reaches 1.188 ($\Delta$NSS$_\text{prior}$\,=\,$+0.462$).
The FiLM advantage over one-hot grows from $+0.013$ under standard
evaluation to $+0.117$ after prior subtraction.
An additive one-hot bias shifts all bottleneck channels uniformly in a
class-specific direction; FiLM's multiplicative $\gamma$ pathway independently
scales each channel, allowing it to amplify channels that encode crowd-specific
features while suppressing channels that encode the centre prior.
This pattern is consistent with feature-wise scaling being less dependent
on the shared centre prior than an additive class bias.

\mypar{Soft FiLM: self-derived conditioning at inference.}
Rather than conditioning on GT density labels, the Soft FiLM variant derives
its conditioning signal from the model's own density head via
$\tilde{\mathbf{e}} = \text{softmax}(\hat{\mathbf{z}}_\text{det}) \mathbf{W}_\text{emb}$,
where $\hat{\mathbf{z}}_\text{det}$ are stop-gradient logits and
$\mathbf{W}_\text{emb}$ is the embedding weight matrix.
This enables continuous interpolation between the three density embeddings
and removes the GT label requirement at inference.
Soft FiLM achieves NSS\,1.424, AUC-J\,0.820, CC\,0.514
($\Delta$NSS$_\text{prior}$\,=\,$+0.446$), only $0.016$ NSS below the
oracle-label \ourmodel{} ($+0.462$ after prior subtraction).  The density
head is trained with category supervision, but no category label is needed
at inference.  This makes Soft FiLM a practical option when CrowdFix labels
are unavailable at deployment.

\mypar{Multiscale FiLM does not improve over bottleneck conditioning.}
Extending density conditioning to all four decoder stages (strides 32, 16, 8, 4)
with independent per-scale FiLM projections (zero-initialised to preserve
training stability) yields NSS\,1.412, AUC-J\,0.819
($\Delta$NSS$_\text{prior}$\,=\,$+0.405$), weaker than both
single-bottleneck variants.  Among the tested locations, the bottleneck
therefore provides the strongest result without additional per-scale
parameters.

\mypar{Continuous VGG16 conditioning: highest standard NSS.}
Replacing the discrete 3-class embedding with a continuous density descriptor derived
from the frozen VGG16 frontend of CSRNet~\citep{csrnet} (conv1-conv4\_3, ImageNet-pretrained)
followed by dilated convolutional backend layers yields NSS\,1.441, AUC-J\,0.821,
CC\,0.520 ($\Delta$NSS$_\text{prior}$\,=\,$+0.416$), the highest
\emph{standard} NSS of all variants, but below the discrete FiLM model's
prior-subtracted gain ($+0.462$).  One possible explanation is that VGG16
features capture scene-level properties that align with the shared spatial
prior, while the explicitly supervised discrete embedding provides cleaner
density-category separation.  The differing standard and prior-subtracted
results suggest that the two conditioning signals may be complementary.

\mypar{DHF1K pretraining neither helps nor hurts.}
Phase~3 asks whether pretraining the VideoSwin encoder on a general-purpose
video saliency dataset (DHF1K; 600 training videos)
before fine-tuning as DensFiLM on CrowdFix provides any benefit.
The DHF1K-pretrained variant achieves
AUC-J\,0.821, NSS\,1.436, CC\,0.520, effectively identical to direct CrowdFix
training from Kinetics-400 initialisation (NSS\,1.437, $\Delta$NSS\,$\leq$\,0.001).
The Kinetics-400-pretrained
Swin3D-S encoder already captures temporally coherent spatiotemporal features
at a much larger scale than DHF1K.  In this setup, the additional DHF1K
pretraining produces no measurable improvement on CrowdFix, whereas the
FiLM layers train successfully from random initialisation.

\mypar{RAFT optical flow does not improve DensFiLM.}
Appending RAFT~\citep{raft} dense optical flow as a second input stream,
fused with the Swin3D bottleneck via cross-attention, yields metrics
identical to \ourmodel{} across all three metrics (AUC-J\,0.820,
NSS\,1.437, CC\,0.517; both seed~42).  Thus, explicit flow provides no
measurable benefit in this configuration.  This result suggests that the
Video Swin features already capture the motion information needed for the
task, or that the available training data are insufficient to exploit the
additional stream.

\mypar{The combined higher-capacity variant returns to baseline.}
The three-branch architecture combines a parallel UpBlock3d temporal decoder
branch, a per-category spatial social force prior, and a learned
gate that fuses the FiLM and 3D branches. It achieves AUC-J\,0.808,
NSS\,1.313, CC\,0.477.  The three-branch NSS of $\approx$1.313 is the most striking result in
the ablation: it is effectively identical to the VideoSwin baseline
($\Delta$NSS\,$\approx$\,0), meaning the additional architecture not only
fails to improve over the FiLM model but also cancels the FiLM gain
entirely.  This outcome is consistent with a bias-variance tradeoff at the
available training scale, although the experiment does not isolate the
contribution of each added component.  The result shows only that this
combined higher-capacity design is not supported by the available CrowdFix
training data.

\subsection{Per-Density-Category Performance}%
\label{sec:per_cat}

Table~\ref{tab:per_cat} breaks down \ourmodel{} performance by density
category, providing insight into how density conditioning benefits each
regime.

\begin{table}[!t]
  \def\arraystretch{1.1}
  \centering
  \footnotesize
  \caption{%
    Per-density-category test performance of \ourmodel{} (seed~42).
    Overall row corresponds to the seed~42 run; Table~\ref{tab:sota}
    reports multi-seed mean\,$\pm$\,std.
    $n$\,=\,number of evaluated test frames.
  }
  \begin{tabular}{lccc}
    \toprule
    \textsc{Category} ($n$)
      & \textsc{AUC-J}$\uparrow$
      & \textsc{NSS}$\uparrow$
      & \textsc{CC}$\uparrow$ \\
    \midrule
    SP - Sparse Pedestrian (1443)  & 0.821 & 1.439 & 0.523 \\
    DF - Dense Free-flowing (1924) & 0.818 & 1.422 & 0.514 \\
    DC - Dense Congested (1989)    & 0.822 & 1.449 & 0.517 \\
    \midrule
    \textit{Overall} (5356)         & 0.820 & 1.437 & 0.517 \\
    \bottomrule
  \end{tabular}%
  \label{tab:per_cat}
\end{table}

DC achieves the highest NSS (1.449), although the differences between
categories are small.  One possible interpretation is that the density label
is particularly useful when individual motion cues are weak or ambiguous.
The parameter analysis below is consistent with category-specific channel
reweighting, but it does not identify the visual semantics of individual
channels.

SP achieves the highest CC (0.523).  This may reflect the more localised
fixation distributions observed in sparse scenes, where individual people
often provide distinct targets.

The SP category spans a wide range of group sizes, from single pedestrians
to groups of 10-20 persons in open plazas.  Fixation strategy is expected
to vary within SP accordingly: individual trajectory tracking for lone
figures, social gaze-following for small groups, and broad scene scanning
for larger gatherings. Yet the per-category NSS above does not resolve
this variation.  Future work could disaggregate SP performance by
approximate group size.

DF achieves the lowest scores across all metrics, consistent with the
inherent difficulty of the free-flowing regime: ground-truth fixations are
broadly distributed over multiple equivalently-attractive individuals, yielding
a diffuse GT saliency map with no dominant peak.  Predicting diffuse
distributions with high NSS may be harder because normalised saliency is
distributed across more plausible locations.  This suggests
that DF may require finer temporal modelling of specific flow-trajectory
emergence points to achieve the per-fixation precision that NSS rewards.

\subsection{FiLM Parameter Analysis}%
\label{sec:film_analysis}

A potential concern is that with only $K\!=\!3$ density classes, the
$3\!\times\!64$ embedding table degenerates to three fixed $(\gamma,\beta)$
vectors, behaving as a class-conditional bias lookup rather than dynamic feature
modulation.  We verify whether the learned parameters encode genuinely
distinct structure by extracting the three concatenated
$[\gamma;\beta]\!\in\!\mathbb{R}^{1536}$ vectors from the trained checkpoint
and computing pairwise cosine similarity between the $\gamma$ components.

The $\gamma$ vectors are nearly orthogonal, with pairwise cosine similarities
of $-0.03$ (SP vs.\ DF), $-0.10$ (SP vs.\ DC), and $-0.06$ (DF vs.\ DC).
These values show that the three classes learn distinct scaling directions
rather than a common offset.  The weakly negative similarities indicate
slight opposing alignment, not strong anti-correlation.

A channel-level inspection shows the asymmetry: the single most
discriminative channel (Ch\,437) exhibits $\gamma_\mathrm{SP}\!=\!+1.09$
vs.\ $\gamma_\mathrm{DC}\!=\!-1.88$ (range\,2.97), while channel 291
shows the complementary pattern ($\gamma_\mathrm{SP}\!=\!-2.54$,
$\gamma_\mathrm{DC}\!=\!+0.14$).  The L2 norms of the full $(\gamma,\beta)$
vectors are comparable across classes (SP\,25.3, DF\,25.9, DC\,24.9),
showing similar modulation magnitudes with distinct directions.

\subsection{Qualitative Analysis}

\begin{figure*}[p]
  \centering
  \setlength{\tabcolsep}{3pt}
  \begin{tabular}{c@{\hspace{4pt}}ccc}
    & \textbf{Best case} & \textbf{Average case} & \textbf{Challenging case} \\[4pt]
    \rotatebox{90}{\parbox{2.8cm}{\centering\textbf{SP}\\\small\textit{Sparse Ped.}}} &
      \includegraphics[width=0.305\textwidth]{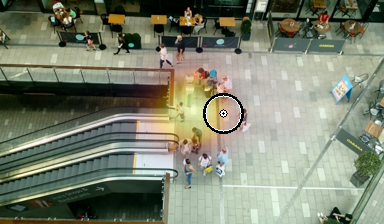} &
      \includegraphics[width=0.305\textwidth]{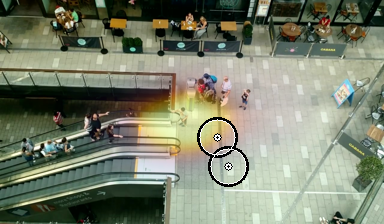} &
      \includegraphics[width=0.305\textwidth]{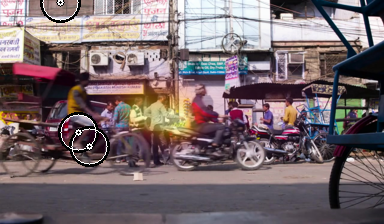} \\
    \multicolumn{1}{c}{} &
      \small v212\,f43 \ NSS\,=\,3.83 &
      \small v163\,f42 \ NSS\,=\,1.44 &
      \small v260\,f81 \ NSS\,=\,$-$0.20 \\[6pt]
    \rotatebox{90}{\parbox{2.8cm}{\centering\textbf{DF}\\\small\textit{Dense Free-flow}}} &
      \includegraphics[width=0.305\textwidth]{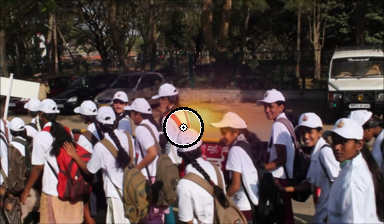} &
      \includegraphics[width=0.305\textwidth]{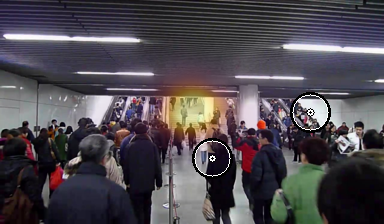} &
      \includegraphics[width=0.305\textwidth]{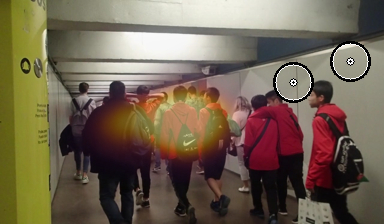} \\
    \multicolumn{1}{c}{} &
      \small v152\,f36 \ NSS\,=\,3.92 &
      \small v064\,f80 \ NSS\,=\,1.43 &
      \small v093\,f17 \ NSS\,=\,$-$0.43 \\[6pt]
    \rotatebox{90}{\parbox{2.8cm}{\centering\textbf{DC}\\\small\textit{Dense Congested}}} &
      \includegraphics[width=0.305\textwidth]{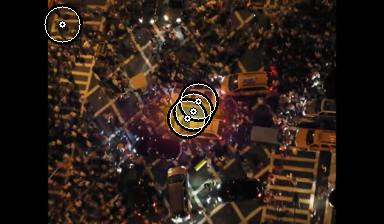} &
      \includegraphics[width=0.305\textwidth]{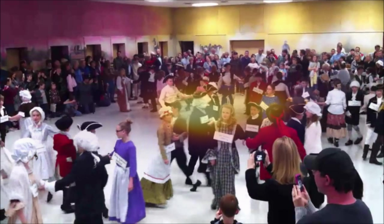} &
      \includegraphics[width=0.305\textwidth]{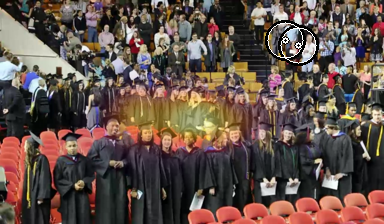} \\
    \multicolumn{1}{c}{} &
      \small v335\,f50 \ NSS\,=\,4.34 &
      \small v160\,f14 \ NSS\,=\,1.47 &
      \small v180\,f38 \ NSS\,=\,$-$0.40 \\
  \end{tabular}
  \caption{%
    Qualitative results stratified by performance tier (nine unique videos).
    Each image overlays the predicted saliency (INFERNO colormap,
    saliency-proportional alpha) and per-observer fixation rings
    (2-degree visual-angle radius; white ring with black halo;
    white centre dot).
    \textbf{Best-case (left):} predicted saliency peaks co-localise
    tightly with observer fixations: spot-attention on individuals (SP),
    diffuse flow coverage (DF), and landmark-level precision (DC).
    \textbf{Average-case (centre):} predictions partially overlap fixation
    clusters with broader spatial spread, consistent with
    above-chance localisation at median performance.
    \textbf{Challenging (right):} negative NSS indicates misplaced
    saliency: the model predicts a central peak while fixations are
    off-centre (an eccentric individual in SP, a lateral flow structure
    in DF, an off-centre landmark in DC), exposing residual centre bias
    in the model predictions.
  }%
  \label{fig:qualitative}
\end{figure*}

Figure~\ref{fig:qualitative} presents a stratified sample of model
behaviour across each density category: one best-case frame (top NSS
in the test split), one average-case frame (NSS near the per-category
median of $\approx$1.43), and one challenging frame (negative NSS,
indicating saliency placed away from observer fixations).  All nine
frames are from distinct videos.  Each image combines the original
frame, the predicted saliency overlay (INFERNO colormap with
saliency-proportional alpha), and per-observer fixation rings (white
ring with black halo at 2-degree visual-angle radius; white centre dot).

\mypar{Best-case frames (left column, NSS $\approx$ 3.8-4.3).}
In the selected best-case examples, the predictions align with the dominant
fixation patterns: spot attention on proximal individuals (SP video~212,
frame~43, NSS\,=\,3.83); diffuse coverage over the pedestrian flow
(DF video~152, frame~36, NSS\,=\,3.92); and precise co-localisation
with a high-contrast scene landmark rather than crowd mass
(DC video~335, frame~50, NSS\,=\,4.34).  These frames share strong
centre alignment between fixation clusters and predicted peaks, so their
high scores likely reflect both content localisation and dataset centre bias.

\mypar{Average-case frames (middle column, NSS $\approx$ 1.4).}
Performance near the per-category mean reveals a broader, less
precise saliency distribution.  Predicted peaks partially overlap
the fixation cluster but with more spatial spread, consistent with
the model averaging over plausible fixation locations rather than
committing to the correct target.  The fixation rings are still
largely enclosed by the saliency blob, consistent with above-chance
localisation at the cost of precision.

\mypar{Challenging frames (right column, NSS $<$ 0).}
Negative NSS indicates that predicted saliency is low at observer fixation
locations.  In all three selected cases the model predicts a central peak while the
observer fixations are displaced to the periphery: an off-centre
individual (SP video~260, frame~81), a lateral flow structure
(DF video~093, frame~17), or an off-centre landmark
(DC video~180, frame~38).  These examples illustrate residual centre bias,
but do not by themselves establish its source.

\section{Discussion}%
\label{sec:discussion}

\subsection{Interpretation and Implications}

The results support two main interpretations.  First, the location of density
conditioning appears more important than adding generic model capacity.  The
one-hot ablation recovers most of the improvement by injecting density at the
semantic bottleneck, while FiLM provides a smaller additional gain through
channel-wise scaling.  This suggests that crowd density is most useful as a
compact signal for selecting among abstract visual features before spatial
decoding.

Second, the centre-prior-subtraction diagnostic increases the measured
conditioning gain from $0.124$ to $0.462$ NSS.  Although this diagnostic is
not a standard benchmark metric, the relative change suggests that the FiLM
model is less dependent on the shared spatial prior than the unconditioned
backbone.  The capacity ablations qualify this conclusion: explicit RAFT flow
and the tested temporal and social-force extension provide no improvement in
the present configuration.  These experiments do not establish that such
components are generally ineffective, only that the available CrowdFix
training data do not support the tested implementations.

\subsection{Limitations and Future Work}

Three limitations bound the scope of this work.
First, \textbf{density taxonomy}: at inference the model can use
either the ground-truth CrowdFix category label or its own auxiliary
density head prediction.  Empirically, the predicted-label variant
achieves NSS\,1.438 vs.\ oracle NSS\,1.437 on seed~42 ($\Delta\!=\!0.001$;
see Table~\ref{tab:sota}), so oracle access is not required in this
evaluation.  However, the SP/DF/DC taxonomy is
CrowdFix-specific; generalising to an unlabelled corpus would require
either a compatible density classifier or unsupervised cluster assignment.
Second, \textbf{single-dataset scope}: all results are on CrowdFix.
Cross-dataset transfer via DHF1K pretraining yields no measurable gain
(Section~\ref{sec:ablation}); whether the conditioning mechanism generalises
to other datasets and labelling schemes remains untested.
Third, \textbf{centre bias and diagnostic evaluation}: the CrowdFix fixation
maps and model predictions both exhibit a centre tendency
(negative-NSS examples in Figure~\ref{fig:qualitative} are all off-centre
misses).  Standard metrics therefore conflate crowd-content-responsive
predictions with centre-prior alignment.  Our prior-subtraction analysis is
a diagnostic rather than a standard benchmark protocol and can produce
negative prediction values.  It should therefore be interpreted as a
relative sensitivity analysis, not as a replacement evaluation metric.

Future work should test the approach on additional datasets and density
taxonomies, investigate larger unlabelled crowd-video pretraining corpora,
and examine whether continuous scene descriptors complement discrete
density conditioning.

\section{Conclusion}%
\label{sec:conclusion}

We introduced \ourmodel, a Video Swin Transformer with lightweight
density-conditioned FiLM modulation at the encoder bottleneck.  The
$\approx$100K-parameter module improves NSS by $9.2\%$ and CC by $8.6\%$
over the unconditioned backbone, yielding the best reported CrowdFix results
we are aware of (AUC-J\,$0.820\pm0.001$, NSS\,$1.434\pm0.007$,
CC\,$0.517\pm0.002$ over four seeds).  The model's own density prediction
matches oracle-label performance at inference.  Overall, the results show
that targeted bottleneck conditioning is a more effective use of limited
CrowdFix training data than the higher-capacity alternatives tested here.

\section*{Acknowledgements}

The author acknowledges CSC - IT Center for Science, Finland, for
computational resources on the LUMI HPC cluster under project~462000131.

\bibliographystyle{unsrtnat}
\bibliography{references}

\end{document}